\documentclass[10pt,twocolumn,letterpaper]{article}

\usepackage[pagenumbers]{cvpr}
\usepackage{multirow} 
\usepackage{graphicx} 
\usepackage{xcolor} 

\definecolor{flowcolor}{HTML}{FF3B30}   
\definecolor{ecolor}{HTML}{FF9500}      
\definecolor{rcolor}{HTML}{FFD60A}      
\definecolor{dancecolor}{HTML}{007AFF}  

\definecolor{cvprblue}{rgb}{0.21,0.49,0.74}
\usepackage[pagebackref,breaklinks,colorlinks,allcolors=cvprblue]{hyperref}
\usepackage{caption} 

\title{
  \textbf{
    \textcolor{flowcolor}{Flow}\textcolor{ecolor}{e}\textcolor{rcolor}{r}\textcolor{dancecolor}{Dance: }
    Mean\textcolor{flowcolor}{Flow} for 
    \textcolor{ecolor}{E}fficient and 
    \textcolor{rcolor}{R}efined 3D \textcolor{dancecolor}{Dance} Generation
  }
}

\author{
\textbf{Kaixing Yang}\textsuperscript{1}\thanks{Equal contribution.} \quad
\textbf{Xulong Tang}\textsuperscript{4}\footnotemark[1] \quad
\textbf{Ziqiao Peng}\textsuperscript{1}\footnotemark[1] \quad
\textbf{Xiangyue Zhang}\textsuperscript{3}
\\[-0.05em]
\textbf{Puwei Wang}\textsuperscript{1}\thanks{Corresponding authors.} \quad
\textbf{Hongyan Liu}\textsuperscript{1}\footnotemark[2] \quad
\textbf{Jun He}\textsuperscript{2}\footnotemark[2]
\\[-0.05em]
\textsuperscript{1}Renmin University of China \quad
\textsuperscript{2}Tsinghua University \quad
\textsuperscript{3}Wuhan University \quad
\textsuperscript{4}Malou Tech Inc
\\[-1.5em]
}

\begin{document}

\twocolumn[{
    \renewcommand\twocolumn[1][]{#1}%
    \maketitle
    \begin{center}
        \centering
        \includegraphics[width=0.95\textwidth]{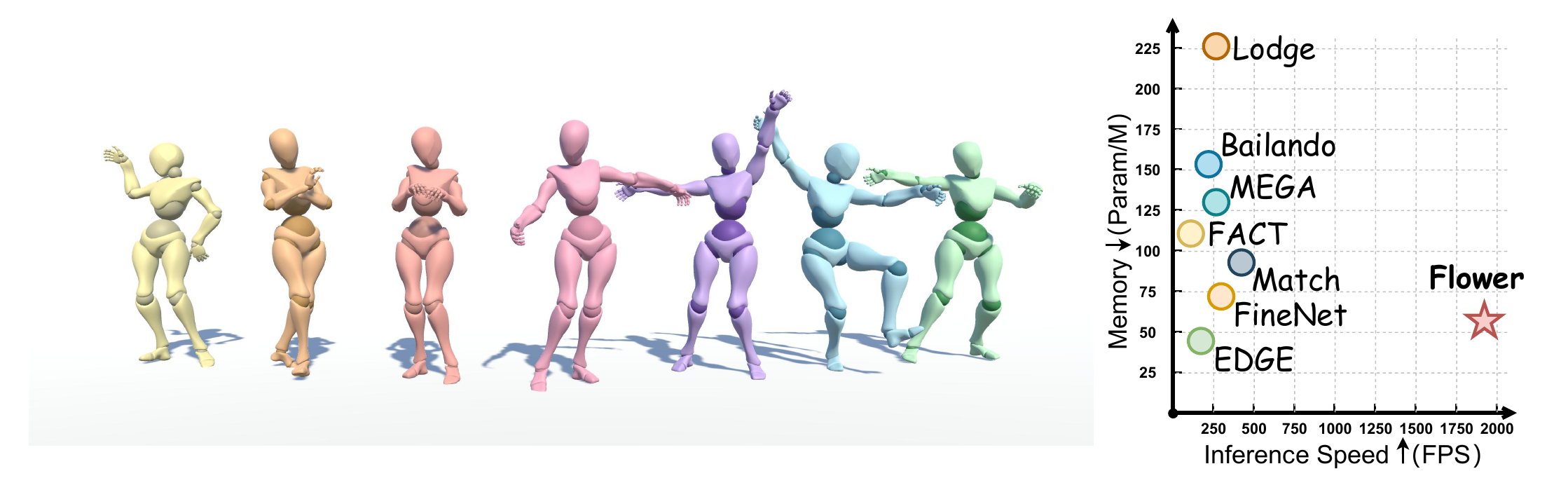}
        \vspace{-0.2in}
        \captionof{figure}{FlowerDance not only generates refined motion with physical plausibility and artistic expressiveness, but also achieves generation efficiency that surpasses that of state-of-the-art (SOTA) methods in inference speed and memory utilization.}
    \vspace{-0.05in}
    \label{fig: teaser}
    \end{center}
}]

\begin{abstract}
Music-to-dance generation aims to translate auditory signals into expressive human motion, with broad applications in virtual reality, choreography, and digital entertainment. Despite promising progress, the limited generation efficiency of existing methods leaves insufficient computational headroom for high-fidelity 3D rendering, thereby constraining the expressiveness of 3D characters during real-world applications. Thus, we propose FlowerDance, which not only generates refined motion with physical plausibility and artistic expressiveness, but also achieves significant generation efficiency on inference speed and memory utilization. Specifically, FlowerDance combines MeanFlow with Physical Consistency Constraints, which enables high-quality motion generation with only a few sampling steps. Moreover, FlowerDance leverages a simple but efficient model architecture with BiMamba-based backbone and Channel-Level Cross-Modal Fusion, which generates dance with efficient non-autoregressive manner. Meanwhile, FlowerDance supports motion editing, enabling users to interactively refine dance sequences. Extensive experiments on AIST++ and FineDance show that FlowerDance achieves state-of-the-art results in both motion quality and generation efficiency. Code will be released upon acceptance. Project page: \url{https://sun-happy-ykx.github.io/FlowerDance/}.

\end{abstract}

\section{Introduction}
Music-to-dance generation is a crucial task that translates auditory input into dynamic motion, with significant applications in virtual reality, choreography, and digital entertainment\cite{li2021ai,li2023finedance}. By automating this process, it enables deeper exploration of the intrinsic relationship between music and movement\cite{yang2024beatdance}, while expanding possibilities for creative content generation. Due to its broad impact, music-to-dance (M2D) generation  has attracted increasing attention\cite{siyao2022bailando,siyao2023bailando++,li2024lodge++}. A broadly applicable M2D system should excel in two aspects: \textbf{(1) Dance Quality}—producing physically plausible, artistically expressive motion; and \textbf{(2) Generation Efficiency}—achieving high inference speed with a low memory utilization.

A substantial body of Music-to-Dance methods has emerged, spanning GANs, autoregressive models, and diffusion approaches. GAN-based methods~\cite{yang2024cohedancers,sun2019deep,huang2021choreography} enhance perceptual naturalness but often degenerate into overly simple, repetitive motion. Autoregressive models ~\cite{siyao2024duolando,yang2025megadance,siyao2022bailando,siyao2023bailando++} incur substantial inference latency due to strictly sequential decoding. Diffusion-based methods~\cite{tseng2023edge,li2023finedance,li2024lodge} achieve strong fidelity but typically require many denoising steps during inference time—attributable to their curved sampling trajectories—resulting in heavy computational overhead. In conclusion, even prior works~\cite{li2024lodge,tseng2023edge,yang2025megadance} have achieved strong dance quality, their limited generation efficiency leaves insufficient computational headroom for high-fidelity 3D rendering, thereby constraining the expressiveness of 3D characters during real-world interactive applications. To this end, we present FlowerDance, delivering excellent dance quality and substantially surpassing prior art in generation efficiency, as shown in Fig. \ref{fig: teaser}.

For generative strategy, FlowerDance integrates MeanFlow with a Physical Consistency Constraint and supports effective Motion Editing. Firstly, from a choreography standpoint, dance creation is a step-wise sculpting process guided by the music: choreographers first initialize a simple and high-entropy motion seed aligned to the music, then progressively refine speed, center-of-mass shifts, and transitions so the movement takes shape smoothly over steps~\cite{butterworth2004teaching,morris2009dance}. Flow Matching~\cite{lipman2022flow,liu2022flow} mirrors this workflow by learning a music-conditioned continuous-time velocity field and transporting a simple prior, step by step, into a complete motion trajectory that is physically plausible and artistic expressive. Furthermore, MeanFlow explicitly models the target distribution’s mean trajectory, aligning the training objective with the inference procedure and enabling high-quality dance generation in just a few steps, substantially improving efficiency. Secondly, we impose a physical consistency constraint that pulls trajectories toward the human motion manifold at each training iteration: we learn the mean velocity from any timestep to the data distribution and apply reconstruction, velocity, and 3D joint position losses on the recovered motion. Thirdly, FlowerDance supports motion editing without additional model training, similar to diffusion-based methods~\cite{tseng2023edge,liu2025gcdance}; to stabilize constrained regions under few-step sampling, we introduce a time-decayed soft masking strategy.

For model architecture, FlowerDance leverages Bidirectional Mamba (BiMamba) backbone with Channel-Level Cross-Modal Fusion. Firstly, BiMamba’s intrinsic sequential inductive bias naturally aligns with the inherent Bidirectional local dependencies between music and dance, making it a natural choice as the backbone for music‑to‑dance generation. Secondly, the Mamba has only $\mathcal{O}(n)$ computational complexity compared to the $\mathcal{O}(n^2)$ complexity of attention‑based Transformers~\cite{vaswani2017attention}. Furthermore, BiMamba supports non‑autoregressive generation, which not only further improves efficiency but also mitigates the exposure‑bias problem in autoregressive methods~\cite{siyao2022bailando,siyao2023bailando++,yang2025megadance} and alleviates transition artifacts at segment boundaries in inpainting‑based approaches~\cite{tseng2023edge,liu2025gcdance}. Different from Text2Motion~\cite{guo2022tm2t,guo2022generating} or Text2Image~\cite{esser2024scaling,ki2024float} tasks, music and dance are inherently temporal aligned. Thus, we introduce a Channel-Level Cross-modal Fusion mechanism in place of conventional cross-attention~\cite{siyao2022bailando,tseng2023edge}. This parameter-free method not only reduces the memory utilization and improves inference speed, but also offers stronger generalization in a non-autoregressive generation setting.

In summary, our contributions in the music-to-dance generation task are as follows:
\begin{itemize}
    \item We propose an efficient and refined framework, named FlowerDance, achieving state-of-the-art (SOTA) performance in both dance quality and generation efficiency on AIST++~\cite{li2021ai} and FineDance~\cite{li2023finedance}.
    \item We develop a novel generative strategy combining MeanFlow with Physical Consistency Constraints, which enables high-quality motion generation with only a few sampling steps, while also supporting effective motion editing for interactive user control.
    \item We design a simple and efficient model architecture with a BiMamba-based backbone and Channel-Level Cross-Modal Fusion, which generates dance in an efficient non-autoregressive manner.
\end{itemize}

\section{Related Work}
\subsection{Flow Matching Generative Models}
Flow Matching (FM)~\cite{lipman2022flow} learns the conditional vector field that transports noise samples to data samples, and performs generation by solving the corresponding ordinary differential equation (ODE). Compared with diffusion-based models such as DDPM~\cite{ho2020denoising}, which rely on stochastic score estimation and long sampling chains, FM offers a simpler and more stable training objective, while enabling deterministic sampling through ODE integration. This makes FM theoretically appealing for efficient generation. To further improve efficiency, several variants of FM have been proposed for few-step sampling. Rectified Flow (RF)~\cite{liu2022flow} constrains the learned ODE to follow straight paths between noise and data, thereby reducing the transport cost. Consistency Flow Matching (CFM)~\cite{yang2024consistency} enforces step-wise consistency so that generation can be carried out in only a few steps. Gaussian Mixture Flow (GMFlow)~\cite{chen2025gaussian} models the transport vector field as a mixture of Gaussian components, allowing more flexible trajectories. Despite these efforts, it remains difficult to achieve high-quality generation with very few steps on complex tasks. To address this, MeanFlow~\cite{geng2025mean} predicts the interval-averaged velocity rather than instantaneous ones, aligning the training objective with the ODE-based inference procedure, thereby enabling high-fidelity dance generation with significantly fewer sampling steps. 

Beyond theoretical advances, FM has also shown strong potential in various generative tasks. In image generation, Stable Diffusion3~\cite{esser2024scaling} incorporates rectified flow into large-scale training, significantly improving efficiency and scalability.
In video generation, FM has been applied with rectified formulations to accelerate frame synthesis~\cite{davtyan2023efficient,ki2024float}, and Pyramid Flow Matching~\cite{jin2024pyramidal} demonstrates coherent temporal modeling with fewer sampling steps.
In audio generation, Voicebox~\cite{le2023voicebox} and Audiobox~\cite{vyas2023audiobox} adopt FM to build large-scale speech and audio models, while Matcha-TTS~\cite{mehta2024matcha} and VoiceFlow~\cite{guo2024voiceflow} leverage conditional and rectified FM for efficient text-to-speech. Interestingly, choreography can be interpreted as a flow-matching process shaped by the potential induced by music, suggesting that employing Flow Matching for music-to-dance generation is a promising but hitherto unexplored direction.

\subsection{Music-to-Dance Generation}
Music and dance are inherently interconnected, which has driven substantial progress in music-to-dance generation (M2D), with most research focusing on 3D dance generation. Prior studies extract musical features using toolkits such as Librosa~\cite{mcfee2015librosa}, Jukebox~\cite{tseng2023edge}, and MERT~\cite{yang2024codancers}, and then predict human motion represented by SMPL parameters~\cite{loper2023smpl,li2021ai} or 2D/3D body keypoints~\cite{siyao2022bailando}. Existing approaches can be broadly grouped into three paradigms: GAN-based, Autoregressive, and Diffusion-based models.  

\textbf{1) GAN-based models.} Generators synthesize motion from music while discriminators provide adversarial feedback. Examples include CoheDancers~\cite{yang2024cohedancers}, DeepDance~\cite{sun2019deep}, and Choreography cGAN~\cite{huang2021choreography}. These methods are simple but prone to mode collapse and lack explicit motion constraints, often producing monotonous and repetitive movements. \textbf{2) Autoregressive models.} These methods typically adopt a two-stage pipeline, curating choreographic units from motion databases followed by autoregressive modeling of music-conditioned distributions over these units. Since units are derived from real motion, results are biomechanically plausible. Works such as Bailando~\cite{siyao2022bailando}, Bailando++~\cite{siyao2023bailando++}, Duolando~\cite{siyao2024duolando}, and MEGADance~\cite{yang2025megadance} fall into this paradigm. However, reliance on local history hampers global music–motion alignment, and step-by-step inference introduces high latency. \textbf{3) Diffusion-based models.} These methods corrupt motion with noise and train denoising networks to iteratively recover sequences conditioned on music, enabling diverse and temporally coherent dances. Representative works include EDGE~\cite{tseng2023edge}, FineNet~\cite{li2023finedance}, Lodge~\cite{li2024lodge}, Lodge++~\cite{li2024lodge++}, and GCDance~\cite{liu2025gcdance}. While achieving high quality, they require many sampling steps due to curved denoising trajectories, resulting in high inference cost. 

While prior works\cite{li2024lodge,tseng2023edge,yang2025megadance} have achieved strong dance quality, their generation efficiency leaves insufficient computational headroom for high-fidelity 3D rendering, thereby constraining the expressiveness of 3D characters during real-world applications. To this end, we present FlowerDance, which couples a MeanFlow-based generative strategy with a BiMamba-based model architecture, delivering excellent dance quality and generation efficiency.

\section{Methodology}

\begin{figure*}[t]
    \centering    \includegraphics[width=0.96\linewidth]{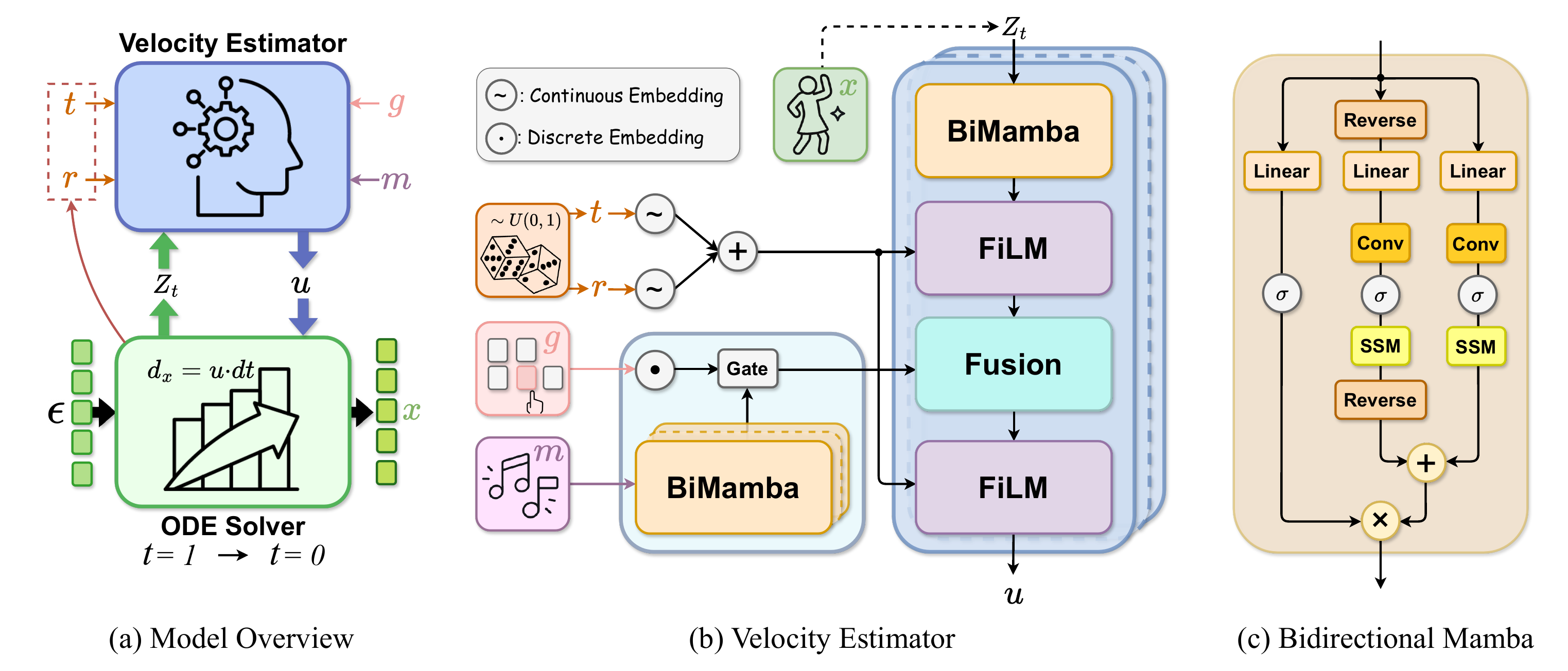}
    \caption{Illustration of model architecture of FlowerDance at different levels.}
    \label{fig:overview}
\end{figure*}

\subsection{Problem Definition}
Given a music sequence $M=\{m_0, m_1, ..., m_T\}$ and a dance genre label $g$, our objective is to synthesize the corresponding dance sequence $D=\{d_0, d_1, ..., d_T\}$, where $m_t$ and $d_t$ denote the music and dance feature at time step $t$. We define each music feature $m_t$ as a 35-dim vector\cite{li2024lodge} extracted by Librosa\cite{mcfee2015librosa}, including 20-dim MFCC, 12-dim Chroma, 1-dim Peak, 1-dim Beat and 1-dim Envelope. We encode genre label $g$ as a one-hot vector. We represent each dance feature as a 147-dim vector $s_t = [\tau; \theta]$, where $\tau$ and $\theta$ encapsulate the root translation and 6-dim rotation representation\cite{zhou2019continuity} of the SMPL\cite{loper2023smpl} model, respectively. Furthermore, we synchronize the music sequence with the dance sequence at a temporal granularity of 30 FPS.

\subsection{Generative Strategy}
\noindent\textbf{MeanFlow.}
In choreographic design, the development of a dance sequence can be viewed as a smooth and steadily paced refinement from an initial unstructured motion toward a coherent target performance, guided consistently by musical rhythm and style\cite{butterworth2004teaching,morris2009dance}. This progression naturally aligns with Flow Matching (FM)~\cite{lipman2022flow}’s assumption that the process can be modeled by an ordinary differential equation (ODE) that transports a standard Gaussian distribution sample $\epsilon \sim \mathcal{N}(0, I)$ to a target distribution sample $x \sim p_{\text{data}}(x)$ by estimating the instantaneous velocity field $v(z_t, t) = \epsilon - x$ along the straight-line flow path $\text{FP}(x, t)$, where time $t$ is randomly sampled from a uniform distribution $U(0,1)$:
\begin{equation}
\begin{aligned}
\text{FP}(x, t) = (1-t)x + t\epsilon
\label{eq: FP}
\end{aligned}
\end{equation}
Here, "straight-line" refers to the standard linear interpolation in latent space between $\epsilon$ and $x$. A vector estimator $v_\theta$ is trained to minimize the ground truth velocity $v(z_t, t)$:
\begin{equation}
\begin{aligned}
\mathcal{L}_{\text{FM}} &= \mathbb{E}\big[\|v_\theta(z_t,t) - v(z_t,t)\|^2\big].
\end{aligned}
\end{equation}
At inference, generation starts from an initial prior sample \(z_{1}\) and iteratively integrates the learned velocity field \(v_\theta(z_t, t)\) until reaching \(z_{0}\) in the data distribution:
\begin{equation}
\frac{d}{dt} z_t = v(z_t, t).
\end{equation}

Even when conditional flows are designed to be straight, latent trajectories in the high-dimensional space are often highly curved. When using large integration intervals, Flow Matching~\cite{lipman2022flow}, which models the instantaneous velocity field, cannot accurately capture these curved trajectories, leading to degraded generation quality. Thus, \textbf{MeanFlow}~\cite{geng2025mean} replaces instantaneous velocity $v(z_t, t)$ with the interval mean velocity $u(z_t, r, t)$, leading to a training objective aligned with the actual ODE inference procedure:
\begin{equation}
u(z_t, r, t) = \frac{1}{t-r} \int_r^t v(z_\tau, \tau)\, d\tau,
\label{eq: meanflow}
\end{equation}
where $(r,t)$ are two time randomly sampled from a uniform distribution $U(0,1)$ with the constraint $t > r$. Differentiating Eq. \ref{eq: meanflow} with respect to $t$ yields the MeanFlow identity:
\begin{equation}
u(z_t, r, t) = v(z_t,t) - (t-r)\frac{d}{dt}u(z_t, r, t).
\end{equation}
A velocity estimator $u_\theta$ is trained to satisfy MeanFlow identity:
\begin{equation}
\begin{aligned}
\mathcal{L}_{\text{MF}}
&= \mathbb{E}\big[\|u_\theta(z_t,r,t) - \text{sg}(u_{\text{tgt}})\|_2^2\big], \\
\text{where} \quad u_{\text{tgt}} &= v(z_t,t) - (t-r)(v(z_t,t) \partial_{z}u_\theta + \partial_t u_\theta)
\end{aligned}
\end{equation}
where $\text{sg}(\cdot)$ denotes stop-gradient. Notably, this reduces to the FM loss when $r=t$. During inference, MeanFlow enables direct mapping by replacing the time integral with the mean velocity. We solve the ODE using an Euler-based discretization scheme, which naturally supports multi-step sampling in a straightforward manner:
\begin{equation}
z_r = z_t - (t-r)u(z_t,r,t).
\label{eq: meanflow-sampling}
\end{equation}

\noindent\textbf{Physical Consistency Constraint.}
Although MeanFlow has achieved notable gains in image~\cite{esser2024scaling,ki2024float} and audio~\cite{vyas2023audiobox,mehta2024matcha} synthesis, 3D motion poses additional challenges due to coupled spatio-temporal dynamics and the need to strictly respect physical plausibility. 
Under the vanilla MeanFlow objective, the model enforces only the mathematical consistency of the velocity field by $\mathcal{L}_{\text{MF}}$, without anchoring trajectories to the human motion manifold. In a high-dimensional, curved, one-to-many conditional mapping, optimization struggles to converge and often drifts toward physically implausible and numerically unstable solutions, manifesting as temporal jitter, global root drift, and abnormal movements.


However, directly imposing conventional physical constraints (e.g., velocity, acceleration, keypoint losses) is not feasible, since the network just predicts interval-averaged velocities instead of original motion. To address this, we additionally sample a time $t_1$ from a uniform distribution $U(0,1)$ in each training iteration. The velocity estimator then directly outputs the mean velocity $u_\theta(z_{t_1}, 0, t_1)$ within the interval $(t_1, 0)$. Subsequently, the motion $\hat{z_0}$ belonging to the target dance distribution is recovered via Eq.~\ref{eq: meanflow-sampling} as $\hat{z_0} = z_{t_1} - t_1 \cdot u_\theta(z_{t_1}, 0, t_1)$. Finally, we incorporate a composite loss function that combines reconstruction, velocity, and 3D joint position losses between the prediction $\hat{z_0}$ and the ground-truth motion $z_0$, ensuring physical consistency and generation stability:

\begin{equation}
\begin{aligned}
\mathcal{L}_{\text{rec}} &= \mathbb{E}\big[\| \hat{z_0} - z_0  \|_2^2 \big], \\
\mathcal{L}_{\text{pos}} &= \mathbb{E}\big[\| FK(\hat{z_0}) - FK(z_0)  \|_2^2 \big], \\
\mathcal{L}_{\text{vel}} &= \mathbb{E}\big[ \| FK(\hat{z_0})' - FK(z_0)'  \|_2^2 \big]
\end{aligned}
\end{equation}
where $FK( \cdot )$ denotes the forward kinematic function that
converts joint angles into joint positions. Our overall training loss is the weighted sum of the MeanFlow objective and the Physical Consistency Constraint, where the weights $\lambda$ are chosen to balance the magnitudes of the losses at the start of training:
\begin{equation}
\mathcal{L} = \lambda_{\text{mf}} \mathcal{L}_{\text{MF}} + \lambda_{\text{rec}} \mathcal{L}_{\text{rec}} + \lambda_{\text{pos}} \mathcal{L}_{\text{pos}} + \lambda_{\text{vel}}  \mathcal{L}_{\text{vel}}.
\end{equation}

\noindent\textbf{Motion Editing.}
Extending the idea of inpainting in diffusion-based motion generation~\cite{liu2025gcdance,tseng2023edge}, FlowerDance also provides a flexible motion editing capability during the sampling stage without additional training cost. This feature enables users to impose a wide range of constraints, such as generating smooth in-between movements in the temporal domain or modifying specific joint trajectories in the spatial domain, as shown in Fig.~\ref{fig:motionedition}. For example, a user may wish to interpolate missing frames between two dance clips or adjust the position of the hands while keeping other joints intact. Since FlowerDance operates with few sampling steps and lacks iterative correction, using a fixed hard mask may cause dynamic mismatches. To address this, we introduce a time-decayed soft mask, where the influence of the known regions gradually diminishes over time, allowing smoother coupling between constrained and unconstrained regions for more coherent inpainting results. 
Given a joint-wise or temporal constraint \(x' \in \mathbb{R}^{N \times 147}\) with positions indicated by a binary mask  \(M \in \{0,1\}^{N \times 147}\) at time $t$, where \(N\) denotes the sequence length, FlowerDance performs constraint-aware denoising as follows:
\begin{equation}
\begin{aligned}
z_{t} := (t \cdot M) \odot \mathrm{FP}(x', t) 
+ (1 - (t \cdot M)) \odot z_{t},
\label{eq:editing_sampling}
\end{aligned}
\end{equation}
where \(\odot\) denotes the Hadamard product, and known regions follow the constraint flow.

\subsection{Model Architecture}
\noindent\textbf{Overview.}
Building on MeanFlow, our model architecture receives as input the time information (start $t$ and end $r$), conditional information (music feature $m$ and genre label $g$), and the current flow‑path state $z_t$, and produces the interval‑averaged velocity $u$, as illustrated in Fig.~\ref{fig:overview}.
For the conditional input, the music feature $m$ is first processed by a multi-layer BiMamba, while the genre label $g$ is encoded via discrete embedding; the resulting genre feature and music feature are then fused using a Gating Mechanism following ~\cite{arevalo2017gated}, dynamically balancing the contributions of music and genre features.
For the time information, the start time $t$ and end time $r$ are first represented as sinusoidal embeddings and then fused via addition. 
Finally, the velocity generation part consists of multi-layer blocks, and each block employs BiMamba for temporal modeling, Feature-wise Linear Modulation (FiLM~\cite{perez2018film}) for integrating time information, and Channel-Level Cross-Modal Fusion (Fusion) for integrating conditional information. 
For long-term generation, FlowerDance produces the entire sequence in a single non-autoregressive pass.
Although simple in design, the synergistic interaction of these components enables FlowerDance to generate high-quality dances with high efficiency.

\begin{table*}[t]
\centering
\renewcommand{\arraystretch}{1.2}
\setlength{\tabcolsep}{5pt}
\caption{Quantitative comparison with SOTAs on the FineDance dataset.}
\label{tab: comparison_finedance}
\scriptsize 
\resizebox{0.95\linewidth}{!}{
\begin{tabular}{l|ccc|cc|c|cc}
\toprule
 & \multicolumn{3}{c|}{\textbf{Quality}} 
 & \multicolumn{2}{c|}{\textbf{Creativity}} 
 & \textbf{Alignment} 
 & \multicolumn{2}{c}{\textbf{Efficiency}} \\ 

\cmidrule(l{-0.5pt}r{0pt}){2-4} 
\cmidrule(l{0pt}r{-0.5pt}){5-6} 
\cmidrule(l{-0.5pt}r{-0.5pt}){7-7} 
\cmidrule(l{0pt}r{0pt}){8-9}

 & FID$_{k}$$\downarrow$ & FID$_{g}$$\downarrow$ & FSR$\downarrow$
 & DIV$_{k}$$\uparrow$ & DIV$_{g}$$\uparrow$
 & BAS$\uparrow$ 
 & Param$\downarrow$ & FPS$\uparrow$ \\ 

\midrule
Ground Truth          
& --     & --     & 0.216
& 9.94 & 7.54 & 0.201
& --    & -- \\

FACT~\cite{li2021ai} [ICCV'21]  
& 113.38 & 97.05 & 0.284
& 3.36  & 6.37 & 0.183
& 120M    & 29 \\ 

MNET~\cite{kim2022brand} [CVPR'22]    
& 104.71 & 90.31 & 0.394
& 3.12  & 6.14 & 0.186
& --    & 26 \\ 

Bailando~\cite{siyao2022bailando} [CVPR'22]      
& 82.81 & 28.17 & 0.188
& 7.74  & 6.25 & 0.202
& 152M    & 188 \\ 

EDGE~\cite{tseng2023edge} [CVPR'23]      
& 94.34 & 50.38 & 0.200
& 8.13  & 6.45 & 0.212
& \textbf{50M}    & 119 \\ 

FineNet~\cite{li2023finedance} [ICCV'23]    
& 65.15 & 23.81 & 0.192
& 5.84  & 5.19 & 0.219
& 94M   & 258 \\ 

Lodge~\cite{li2024lodge} [CVPR'24]      
& 50.00 &  35.52 & \textbf{0.028}
& 5.67  &  4.96 &  0.226
& 235M  & 224 \\ 

MEGA~\cite{yang2025megadance} [NeurIPS'25]
& 50.00 & \textbf{13.02} & 0.243
& 6.23 & 6.27 & 0.226
& 120M & 238 \\ 

Match~\cite{yang2025matchdance} [Arxiv'25]
& 36.68 & 29.47 & 0.328
& 6.45 & 6.98 & 0.227
& 102M & 345 \\ 

FlowerDance (\textbf{Ours})
& \textbf{29.73} & \underline{19.59} & \underline{0.147}
& \textbf{8.42} & \textbf{7.18} & \textbf{0.232}
& \underline{63M} & \textbf{2008} \\ 

\bottomrule
\end{tabular}
}
\vspace{-0.1in}
\end{table*}

\begin{table}[t]
\centering
\renewcommand{\arraystretch}{1.2}
\setlength{\tabcolsep}{5pt}
\caption{Comparison with SOTAs on the AIST++ dataset.}
\label{tab: comparison_aist}
\resizebox{\linewidth}{!}{
\begin{tabular}{l|cc|cc|c}
\toprule
 & FID$_{k}$$\downarrow$ & FID$_{g}$$\downarrow$
 & Div$_{k}$$\uparrow$ & Div$_{g}$$\uparrow$
 & BAS$\uparrow$ \\

\midrule
Ground Truth          
& - & -
& 8.19 & 7.45
& 0.237 \\

DanceNet~\cite{zhuang2022music2dance} [TOMM'22]
& 69.18 & 25.49
& 2.86 & 2.85
& 0.143 \\

FACT~\cite{li2021ai} [ICCV'21]    
& 35.35 & 22.11
& 5.94 & 6.18
& 0.221 \\

Bailando~\cite{siyao2022bailando} [CVPR'22]      
& 28.16 & \textbf{9.62}
& \textbf{7.83} & 6.34
& 0.233 \\

EDGE~\cite{tseng2023edge} [CVPR'23]      
& 42.16 & 22.12
& 3.96 & 4.61
& 0.233 \\

Lodge~\cite{li2024lodge} [CVPR'24] 
& 37.09 & 18.79
& 5.58 & 4.85
& \textbf{0.242} \\

FlowerDance (\textbf{Ours})
& \textbf{20.50} & \underline{15.75}
& \underline{6.95} & \textbf{6.52}
& 0.227 \\

\bottomrule
\end{tabular}
}
\vspace{-0.1in}
\end{table}

\noindent\textbf{Bidirectional Mamba.}
Our choice of Mamba is motivated by the following considerations: (1) Music and dance demand strong local continuity between movements. Transformer is inherently position-invariant and captures sequence order only through positional encodings~\cite{vaswani2017attention}, which limits its deep understanding of local dependencies. Owing to its inherent sequential inductive bias, Mamba~\cite{gu2023mamba} has demonstrated strong performance in modeling fine-grained local dependencies ~\cite{xu2024mambatalk,fu2024mambagesture}. (2) Mamba has only $\mathcal{O}(n)$ computational complexity compared to the $\mathcal{O}(n^2)$ complexity of attention‑based Transformers~\cite{vaswani2017attention}. (3) Mamba supports non‑autoregressive generation, which not only further improves efficiency but also mitigates the exposure‑bias problem in autoregressive methods~\cite{siyao2022bailando,siyao2023bailando++,yang2025megadance} and alleviates transition artifacts at segment boundaries in inpainting‑based approaches~\cite{tseng2023edge,liu2025gcdance}. Furthermore, temporal dependencies in both music and dance are inherently bidirectional, but standard Mamba models sequences only in a single direction. To address this limitation, we introduce Bidirectional Mamba (BiMamba), which captures information from both temporal directions. As shown in Fig.~\ref{fig:overview}, the sequence is processed by forward and backward Mamba branches; their outputs are fused by addition and refined through a multiplicative skip connection, enhancing gradient flow and feature preservation. This bidirectional design improves motion coherence and strengthens music–dance alignment.

Specifically, Selective State Space model (Mamba) incorporates a Selection mechanism and a Scan module (S6)~\cite{gu2023mamba} to dynamically select salient input segments for efficient sequence modeling. Unlike the S4 model~\cite{gu2021efficiently} with fixed parameters $A, B, C$, and scalar $\Delta$, Mamba adaptively learns these parameters via fully-connected layers, enhancing generalization capabilities. Mamba employs structured state-space matrices, imposing constraints that improve computational efficiency. For each time step $t$, the input $x_t$, hidden state $h_t$, and output $y_t$ follow:
\begin{equation}
\begin{aligned}
h_t &= \bar{A}_t h_{t-1} + \bar{B}_t x_t, \\
y_t &= C_t h_t,
\end{aligned}
\end{equation}
where $\bar{A}_t, \bar{B}_t, C_t$ are dynamically updated parameters. Through discretization with sampling interval $\Delta$, the state transitions become:
\begin{equation}
\begin{aligned}
\bar{A} &= \exp(\Delta A), \\
\bar{B} &= (\Delta A)^{-1} (\exp(\Delta A) - I) \cdot \Delta B, \\
h_t &= \bar{A} h_{t-1} + \bar{B} x_t,
\end{aligned}
\end{equation}
where $(\Delta A)^{-1}$ denotes the inverse of $\Delta A$ and $I$ is the identity matrix; the scan module captures temporal dependencies using shared trainable parameters over time.

\noindent\textbf{Channel-Level Cross-Modal Fusion.}
Different from conventional \emph{cross-attention}, we adopt \emph{element-wise addition} for channel-level fusion for several reasons:  
(1) Music and dance are naturally aligned frame-by-frame, and temporal dependencies have already been fully modeled by BiMamba;  
(2) 3D dance datasets are typically small-scale, and the parameter-free addition operation offers stronger generalization in a non-autoregressive generation setting;  
(3) The lightweight approach reduces the parameters and improves efficiency.


\section{Experiment}
\subsection{Dataset}
\noindent\textbf{FineDance.} \textit{FineDance}~\cite{li2023finedance} is the largest public dataset for 3D music-to-dance generation, featuring professionally performed dances captured via optical motion capture. It provides 7.7 hours of motion data at 30 fps across 16 distinct dance genres. Following~\cite{li2024lodge}, we evaluate on test-set music clips, generating 1024-frame (34.13s) dance sequences. 

\noindent\textbf{AIST++.} \textit{AIST++}~\cite{li2021ai} is a widely used benchmark comprising 5.2 hours of 60 fps street dance motion, covering 10 dance genres. Following~\cite{li2021ai}, we use test-set music clips to generate 1200-frame (20.00s) sequences.

\begin{figure*}[t]
    \centering
    \includegraphics[width=\linewidth]{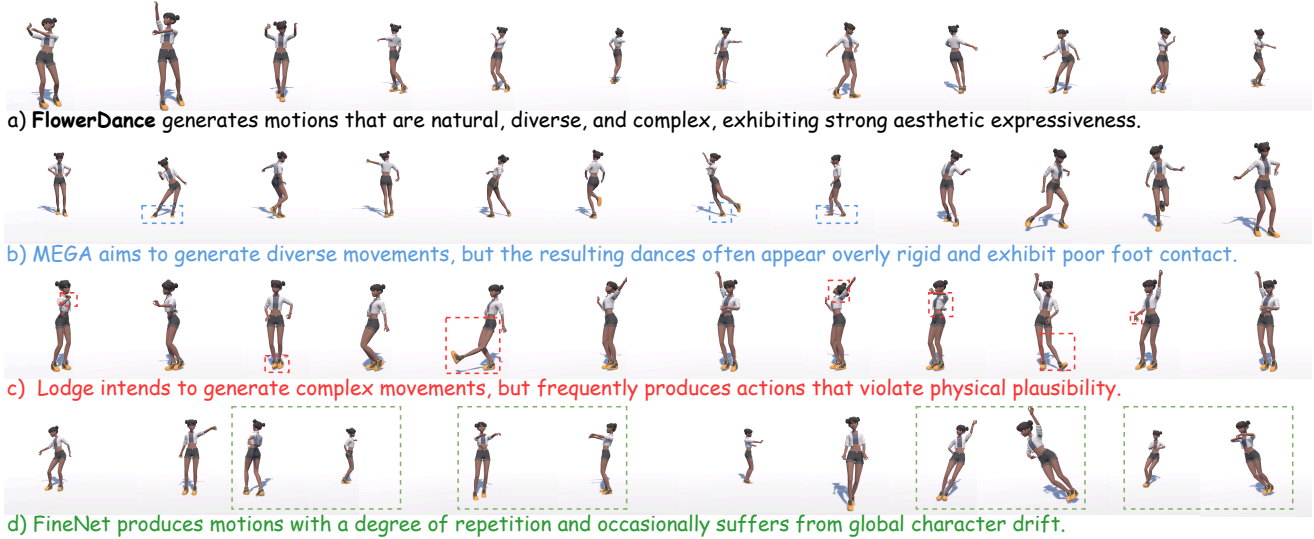}
    \vspace{-0.05in}
    \caption{Qualitative Analysis on a typical Eastern Folk music clip.}
    \label{fig: exp}
    \vspace{-0.1in}
\end{figure*}

\subsection{Quantitative Evaluation}
\noindent\textbf{Dance Quality.}
We evaluate FlowerDance against state-of-the-art (SOTA) baselines on both the FineDance~\cite{li2023finedance} and AIST++~\cite{li2021ai} datasets using a comprehensive suite of metrics. For each generated sequence, we report Fréchet Inception Distance (FID) and Diversity (DIV) across two feature spaces~\cite{li2021ai, siyao2022bailando}: (1) kinetic (k), capturing motion dynamics, and (2) geometric (g), encoding spatial joint relations. To assess physical plausibility, we compute the Foot Slide Ratio (FSR)~\cite{tseng2023edge, li2024lodge}. To measure music–motion synchronization, we utilize the Beat Alignment Score (BAS)~\cite{li2021ai, li2024lodge}. On the FineDance dataset (Tab.~\ref{tab: comparison_finedance}), FlowerDance achieves the best results in $FID_k$ (29.73), $DIV_k$ (8.42), $DIV_g$ (7.18), and BAS (0.232). It also attains competitive performance in $FSR$ (0.147) and $FID_g$ (19.59). On the AIST++ dataset (Tab.~\ref{tab: comparison_aist}), FlowerDance obtains the lowest $FID_k$ (20.50) and the highest $DIV_g$ (6.52). It further shows strong performance in $FID_g$ (15.75), $DIV_k$ (6.95), and BAS (0.227). Taken together, these results establish FlowerDance as a state-of-the-art (SOTA) approach, delivering balanced and competitive performance across metrics and datasets.

\begin{table}[t]
\centering
\renewcommand{\arraystretch}{1.2}
\setlength{\tabcolsep}{5pt}
\caption{User study results on the FineDance dataset.}
\label{tab: user_study}
\tiny
\resizebox{0.7\linewidth}{!}{
\begin{tabular}{l|ccc}
\toprule
 & DQ$\uparrow$ & DS$\uparrow$ & DC$\uparrow$ \\
\midrule

FineNet~\cite{li2023finedance}   
& 3.72 & 3.53 & 3.51 \\

Lodge~\cite{li2024lodge}  
& 4.09 & 3.89 & 4.22 \\

MEGA~\cite{yang2025megadance}
& 4.12 & 4.23 & 4.27 \\

FlowerDance
& \textbf{4.18} & \textbf{4.41} & \textbf{4.33} \\

\bottomrule
\end{tabular}
}
\vspace{-0.1in}
\end{table}

\noindent\textbf{Generation Efficiency.}
As shown in Tab.~\ref{tab: comparison_finedance}, we quantitatively evaluate generation efficiency in terms of memory (model parameters, Param) and inference speed (frames per second, FPS), and report results for generating a 1024-frame (34.13s) sequence. Benefiting from the BiMamba-based backbone and the MeanFlow-based generative strategy, FlowerDance \emph{achieves the highest FPS by a significant margin} while maintaining a competitive Param, only slightly larger than FineNet~\cite{li2023finedance}. This strong efficiency enables seamless integration into real-time interactive systems, where rapid feedback is crucial for user engagement in practice.

\noindent\textbf{User Study.}
Dance’s inherent subjectivity makes user feedback essential for evaluating generated movements~\cite{legrand2009perceiving} in the music-to-dance generation task. Following\cite{yang2025megadance}, we select 40 in-the-wild music segments (34.13s) and generate dance sequences using above models. These sequences are evaluated through a double-blind questionnaire, by 40 participants with backgrounds in dance practice. The questionnaires are based on a 5-point scale (Great, Good, Fair, Bad, Terrible) and assess three aspects: Dance Synchronization ($DS$, alignment with rhythm and style), Dance Quality ($DQ$, biomechanical plausibility and aesthetics), and Dance Creativity ($DC$, originality and range). As shown in Tab.~\ref{tab: user_study}, FlowerDance significantly outperforms all baselines across user-rated metrics (i.e., $DQ =4.18$, $DS = 4.41$, $DC = 4.33$). Its high scores in various aspects underscore its superiority in generating movements in terms of human preferences.

\subsection{Qualitative Evaluation}
To assess the visual quality of the generated dances, we conduct a qualitative comparison between FlowerDance and representative baselines, as shown in Fig.~\ref{fig: exp}. MEGA~\cite{yang2025megadance} seeks to promote diversity, yet its outputs often appear rigid and suffer from poor foot–ground contact. FineNet~\cite{li2023finedance} produces motions with a noticeable degree of repetition and sometimes causes the entire character to drift globally. Lodge~\cite{li2024lodge} attempts to generate complex movements,but its results frequently contain physically implausible actions, such as unnatural bending of the torso and unrealistic limb articulations. In contrast, FlowerDance produces motions that are natural, diverse, and physically coherent, while also demonstrating strong aesthetic expressiveness. This superior performance stems from two main factors: (1) the powerful generative capability of MeanFlow-based model with the Physical Consistency Constraint, enabling the synthesis of complex and diverse movements while ensuring generation stability and physical plausibility; (2) the strong temporal understanding of the BiMamba‑based architecture with Channel‑Level Cross‑Modal Fusion, enabling coherent long‑range choreography while maintaining alignment with musical cues.

\subsection{Ablation Study}
Tab. 4-5 summarize the quantitative statistics, while qualitative visualizations are provided in supplementary video.

\noindent\textbf{1) Generative Model.} 
We compare MeanFlow against Rectified Flow (RectFlow)~\cite{liu2022flow} and Diffusion~\cite{tseng2023edge} under different ODE sampling steps, using the same architecture for fairness. We omit BAS because its unstable jitters can coincidentally align with beats and yield misleadingly high scores. MeanFlow attains state-of-the-art (SOTA) performance with 20 sampling steps, remains near-SOTA at 10 steps, and still produces reasonable results with only 5 steps; performance collapses at a single step, indicating one-step generation is an open but promising direction. From visual inspection, 20-step generations are fully stable, 10-step outputs sometimes show minor jitter, and 5-step outputs occasionally exhibit noticeable jitter. However, RectFlow and Diffusion require 50 steps to reach the quality that MeanFlow achieves with 10 steps. The key reason is that MeanFlow predicts interval-averaged velocity rather than instantaneous velocity, aligning the training objective with the ODE updates used at inference and removing the training–inference mismatch present in RectFlow. Diffusion models rely on progressive denoising to correct errors over many small steps; in few-step regimes, each update must span a large time interval and, without explicit large-step velocity modeling, denoising effectively becomes long-range interpolation, producing trajectory mismatches and loss of fine motion details. Overall, integrating MeanFlow enables FlowerDance to synthesize high-fidelity dances with far fewer sampling steps, thereby significantly improving generation efficiency.

\noindent\textbf{2) Model Backbone.}
To evaluate the effect of the trans-temporal backbone, we compare BiMamba against Mamba, Transformer (II) and Transformer (NAR). Both II and NAR are trained on 8‑second clips; at inference, II iteratively inpaints to 34.13s, whereas NAR generates the full 34.13s in a single pass with non-autoregression.
BiMamba substantially outperforms both alternatives, as it effectively captures bidirectional dependencies between music and dance, achieving superior generation quality. 
Mamba is more efficient but underperforms on metrics and lacks aesthetic appeal and motion diversity. Transformer (NAR) fails to generalize to long sequences—metrics collapse and outputs degenerate into repetitive in-place jittering, owing to self-attention’s scale-dependent positional extrapolation. Transformer (II) attains decent metrics (slightly below BiMamba) but incurs excessive latency; visually, it occasionally exhibits “teleportation,” an inherent boundary limitation of inpainting-based approaches. 
Overall, BiMamba serves as an effective and efficient backbone for FlowerDance.


\begin{table}[t]
\centering
\renewcommand{\arraystretch}{1.2}
\setlength{\tabcolsep}{5pt}
\caption{Effect of the Generative Model under different ODE sampling steps (S) during inference on the FineDance dataset.}
\vspace{-0.05in}
\label{tab:ablation_steps}
\resizebox{0.9\linewidth}{!}{
\scriptsize
\begin{tabular}{l|c|ccc|cc|c}
\toprule
& S & FID$_{k}$$\downarrow$ & FID$_{g}$$\downarrow$ & FSR $\downarrow$
& Div$_{k}$$\uparrow$ & Div$_{g}$$\uparrow$
& FPS $\uparrow$ \\
\midrule
Ground Truth
& - & -- & -- & 0.216
& 9.94 & 7.54
& - \\
\midrule 
\multirow{3}{*}{MeanFlow} 
& 5 & 29.05 & 55.96 & 1.206
& 7.62 & 6.73
& 3531 \\
& 10 & 26.17 & 25.00 & 0.201
& 8.20 & 7.10
& 2725 \\
& 20 & 29.73 & 19.59 & 0.147
& 8.42 & 7.18
& 2008 \\
\midrule 
\multirow{2}{*}{RectFlow} 
& 20 & 49.59 & 39.65 & 0.247
& 8.18 & 7.21
& 2275 \\
& 50 & 33.47 & 38.81 & 0.251
& 7.01 & 6.60
& 1066 \\
\midrule 
\multirow{2}{*}{Diffusion} 
& 20 & 66.46 & 48.25 & 0.274
& 6.78 & 6.38  
& 1462 \\
& 50 & 32.06 & 28.44 & 0.207 
& 8.75 & 7.04
& 890 \\
\bottomrule
\end{tabular}
}
\vspace{-0.1in}
\end{table}

\begin{table}[t]
\centering
\renewcommand{\arraystretch}{1.2}
\setlength{\tabcolsep}{5pt}
\caption{Ablation Study on the FineDance dataset.}
\vspace{-0.05in}
\label{tab:ablation}
\resizebox{\linewidth}{!}{
\scriptsize
\begin{tabular}{l|ccc|cc|c|c}
\toprule
 & FID$_{k}$$\downarrow$ & FID$_{g}$$\downarrow$ & FSR$\downarrow$
 & Div$_{k}$$\uparrow$ & Div$_{g}$$\uparrow$
 & BAS$\uparrow$ & FPS $\uparrow$ \\

\midrule
Ground Truth   
& -- & -- & 0.216 
& 9.94 & 7.54 & 0.201 & -- \\

BiMamba $\to$ Mamba
& 39.40 & 38.93 & 0.197
& 8.69 & 7.07
& 0.219 & \textbf{2387} \\

BiMamba $\to$ Transformer(NAR)
& NaN & NaN & NaN
& NaN & NaN
& NaN & 1829 \\

BiMamba $\to$ Transformer(II)
& \textbf{23.29} & 21.02 & 0.191
& 8.04 & 7.01
& 0.226 & 218 \\

Addition $\to$ Cross Attention
& 32.40 & 24.45 & 0.194
& \textbf{9.12} & 6.90
& 0.229 & 1463 \\

w/o PCC
& NaN & NaN & NaN
& NaN & NaN
& NaN & 2008 \\

FlowerDance (\textbf{Full Model})
& 29.73 & \textbf{19.59} & \textbf{0.147}
& 8.42 & \textbf{7.18} & 0.232 & 2008 \\

\bottomrule
\end{tabular}
}

\end{table}

\noindent\textbf{3) Cross Modal Fusion.}
We further investigate the effect of channel-level cross-modal fusion by replacing the element-wise addition (ADD) with the widely used cross-attention (CA). Overall, ADD shows some improvements over CA. In practice, motions generated by CA are also reasonable, but those produced by ADD tend to align more accurately with musical beats. While the gains are modest, ADD reduces computational cost and inference time, thereby improving FlowerDance’s generation efficiency.

\noindent\textbf{4) Physical Consistency Constraint.}
We analyze the Physical Consistency Constraint (PCC) via ablation. PCC introduces inductive biases that anchor optimization to plausible human kinematics, preventing divergence in early training and improving stability. Moreover, it adds no inference-time computational overhead. Without PCC, training fails to converge—metrics become NaN and outputs collapse into invalid motions with severe temporal jitter and global drift—highlighting PCC’s critical role in FlowerDance.

\begin{figure}[t]
    \centering
    \includegraphics[width=0.95\linewidth]{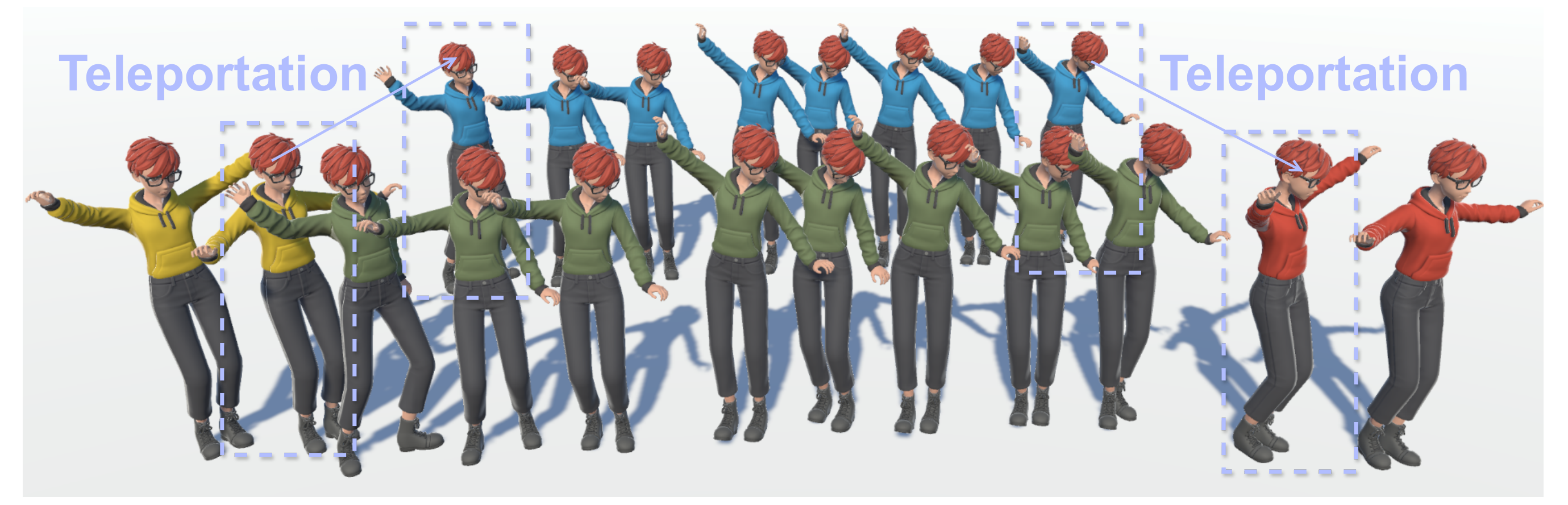}
    \vspace{-0.05in}
    \caption{
    Given start (yellow) and end (red) motions, FlowerDance generates smooth in-between motions with a soft mask (green), whereas a hard mask (blue) causes teleportation artifacts.
    }
    \label{fig:motionedition}
    \vspace{-0.1in}
\end{figure}

\subsection{Qualitative Analysis on Motion Editing}
Motion editing is an important step in dance generation, as it allows users to refine results through interactive control. As shown in Fig.~\ref{fig:motionedition}, FlowerDance with soft mask generates smooth and high-quality in-between motions given start and end clips.  Without the proposed time-decayed soft constraint, FlowerDance (hard mask) can still produce reasonable motions within the inpainted segment, but abrupt discontinuities often occur at the transition boundaries between fixed and inpainted regions, severely compromising the fluency of the generated dance. Details are provided in the supplementary video.

\section{Conclusion}
In this work, we introduce FlowerDance, an efficient and refined framework for music-to-dance generation, achieving state-of-the-art performance (SOTA) in both dance quality and generation efficiency. By incorporating MeanFlow and Physical Consistency Constraints, FlowerDance ensures faithful motion trajectories while requiring only a few sampling steps. Moreover, FlowerDance leverages the BiMamba backbone and Channel-Level Cross-Modal Fusion to construct its model architecture. Meanwhile, FlowerDance also supports motion editing, enabling users to interactively refine dance sequences.

{
    \small
    \bibliographystyle{ieeenat_fullname}
    \bibliography{main}
}


\end{document}